\pdfoutput=1

\documentclass[11pt]{article}

\usepackage[]{ACL2023}

\usepackage{times}
\usepackage{latexsym}
\usepackage{amsfonts}
\usepackage{amssymb}
\usepackage{amsmath}
\usepackage{algorithm2e}
\usepackage{graphicx}
\usepackage{tabularx}
\usepackage{pifont}
\SetKwComment{Comment}{/* }{ */}
\RestyleAlgo{ruled}

\usepackage[T1]{fontenc}

\usepackage[utf8]{inputenc}

\usepackage{microtype}
\usepackage{multirow}

\usepackage{inconsolata}
\usepackage{arydshln}
\usepackage{pifont}
\newcommand{\cmark}{\ding{51}}%
\newcommand{\xmark}{\ding{55}}%
\usepackage{graphics}

\newcolumntype{b}{>{\centering} X}
\newcolumntype{s}{>{\centering\hsize=.2\hsize}X}

\title{MATTER: \underline{M}emory-\underline{A}ugmented \underline{T}ransformer\\ Using He\underline{ter}ogeneous Knowledge Sources}

\author{Dongkyu Lee$^{1}$\Thanks{This work was done while Dongkyu was a summer intern at Amazon AGI}\quad Chandana Satya Prakash$^{2}$\quad Jack FitzGerald$^{2}$\quad Jens Lehmann$^{2}$\\$^{1}$Department of Computer Science and Engineering, HKUST \\$^{2}$Amazon AGI\\\texttt{movingkyu@lgresearch.ai}}

\begin{document}
\maketitle
\begin{abstract}
Leveraging external knowledge is crucial for achieving high performance in knowledge-intensive tasks, such as question answering. The retrieve-and-read approach is widely adopted for integrating external knowledge into a language model. However, this approach suffers from increased computational cost and latency due to the long context length, which grows proportionally with the number of retrieved knowledge. Furthermore, existing retrieval-augmented models typically retrieve information from a single type of knowledge source, limiting their scalability to diverse knowledge sources with varying structures. In this work, we introduce an efficient memory-augmented transformer, called MATTER, designed to retrieve relevant knowledge from multiple heterogeneous knowledge sources. Specifically, our model retrieves and reads from both unstructured sources (paragraphs) and semi-structured sources (QA pairs) in the form of fixed-length neural memories. We demonstrate that our model outperforms existing efficient retrieval-augmented models on popular QA benchmarks in terms of both accuracy and speed. Furthermore, MATTER achieves competitive results compared to conventional retrieve-and-read models while having 100x throughput during inference.
\end{abstract}

\section{Introduction}

Retrieval-augmented models have enhanced performance in various natural language processing tasks, such as open-domain question answering \citep{fid, rag}. These models leverage a two-step process: an initial retrieval phase to gather relevant information from a knowledge source (retrieve), followed by a reading or comprehension phase to generate responses based on the retrieved context and input (read). 

However, the strong performance of retrieval-augmented QA models is offset by a substantial drawback of high inference latency \citep{emat, qamat}. \citep{emat, fid} demonstrate that even with relatively small reader models, such as \texttt{T5-base}, retrieve-and-read QA models struggle to process more than 10 questions per second. 
Recent studies attribute the problem to the increase in context length for a reader to condition on \citep{fido, fid-light, emat}. For instance, the Fusion-in-Decoder model \citep{fid} retrieves 100 documents, each comprising of 250 tokens, resulting in the reader model attending to 25,000 tokens during answer generation; this poses a significant bottleneck during inference. 

To address this limitation, \citep{emat} transforms the retrieved texts as neural memories. A neural memory is an efficient way of storing knowledge with a fixed length latent representation \citep{knn_lm, cai-etal-2021-neural}. As a result, memory-augmented QA models generate an answer conditioned on retrieved neural memories, rather than retrieved raw text. This approach shortens the context length, enabling memory-augmented models to respond to several hundred questions per second \citep{emat}. While this improves the throughput compared to conventional retrieval-augmented models, there is still room for improvement in performance. 

Another critical drawback of existing retrieval-augmented models, both conventional and memory-based approaches, is their narrow focus on a single type of knowledge source, either QA pairs or Wikipedia articles \citep{paq, emat, qamat}.
External knowledge sources come in various formats, such as unstructured, semi-structured, and structured, each with its own merits and use cases. For instance, unstructured data, like a Wikipedia paragraph, is easily accessible and often covers a broad range of topics. However, it may suffer from noise and a lack of precision. In contrast, semi-structured knowledge, such as a question-answer pair (QA), is concise and clear but can be challenging to gather. 
Previous retrieve-and-read approaches have mainly utilized a single type of knowledge source, and this limited focus results in reduced scope and coverage of knowledge.
A potential approach to overcome this limitation is by synchronizing the knowledge structure, such as converting a Wikipedia paragraph into QA pairs using a QA generation pipeline, as demonstrated in \citep{qamat}. However, this format transformation process comes with significant computational cost and the potential risk of introducing noise or corrupting knowledge during the transformation.


To address these limitations, we introduce \textbf{MATTER}, a novel memory-augmented QA model designed to retrieve information from diverse knowledge sources. Unlike existing retrieval models, MATTER retrieves from multiple heterogeneous knowledge sources. This allows our model to maintain a comprehensive and type-agnostic knowledge index, enabling retrieval and conditioning on a broader range of knowledge snippets in response to questions. Moreover, our model cross-encodes a given question and retrieved neural memories, ensuring a comprehensive understanding of input and context.
With this efficient cross-encoding capability and access to heterogeneous knowledge sources, our model significantly outperforms existing efficient QA models in both zero-shot and fine-tuned settings. Furthermore, it achieves a remarkable 100x throughput improvement over raw text-augmented QA models like FiD \citep{fid}, while maintaining competitive performance. Overall, our approach strikes a balance between speed and performance demonstrated over popular QA benchmarks and supported by in-depth analysis.


\section{Related Work}

Question answering is a central task in the NLP community, and various approaches have emerged to address it, falling into three main categories. Firstly, closed-book approaches like \texttt{T5} \citep{t5} and BART \citep{bart} have gained prominence relying solely on the input question and parametric knowledge to generate an answer. These QA models remain suboptimal as they rely exclusively on parametric knowledge \citep{roberts-etal-2020-much}.

The second category comprises retriever-only models, where a retriever fetches a relevant QA pair based on the input question, often with a reranker to enhance QA performance \citep{paq, dpr, two_step}. This approach restricts the knowledge source to QA pairs and relies on strong overlap between the candidate question and QA pairs in the index.

The third category is the retrieve-and-read pipeline models, which have become the standard for building QA models with strong performance \citep{fid, fid_kg, rag}. This method involves a retriever fetching pertinent knowledge, followed by a separate reader model that generates answers based on the acquired context. However, it's worth noting that while these retrieve-and-read models excel in performance and combat hallucination, they are notorious for their slow inference speed \citep{fid-light, fido, emat}.
To address this issue, various strategies have been proposed, such as \citep{DBLP:journals/corr/abs-2112-08560} and \citep{wu-etal-2020-dont}, that dynamically determine which retriever results to read, aiming to reduce computational overhead during inference. Additionally, \citep{fido} eliminates cross-attention from most decoder layers and incorporates multi-query attention. EMAT \citep{emat} takes a novel approach by having the retriever fetch neural memories, significantly accelerating the answer generation process.


\section{Approach}

\subsection{Task Definition}


\begin{figure*}
    \centering
    \includegraphics[width=0.9\textwidth]{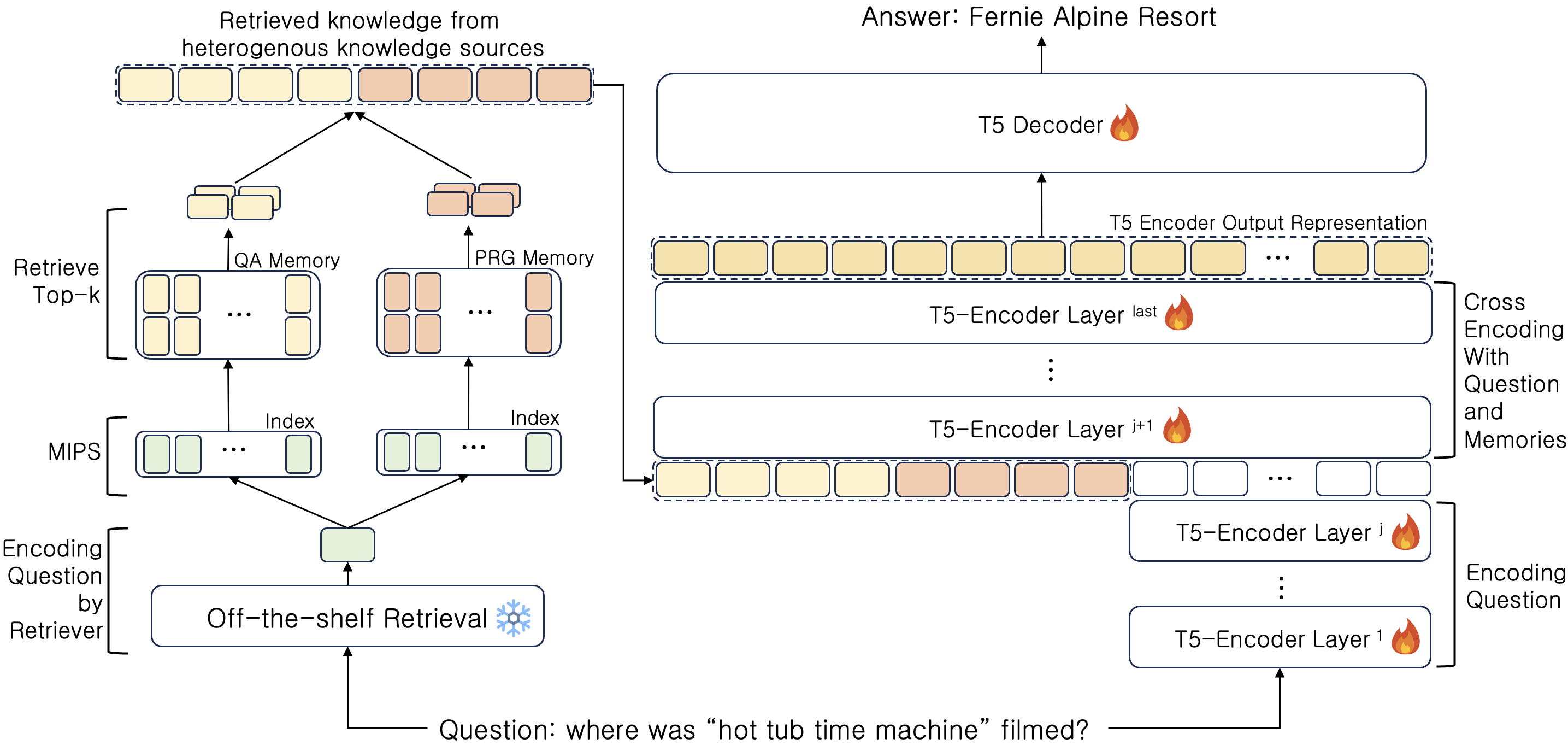}
    \caption{The proposed model with memories retrieved from two different types of knowledge sources: semi-structured (QA) and unstructured (paragraph). In the figure, a memory is represented with two length vectors, and $k$ is set to 4; the model is retrieving a total of 4 memories for inference, 2 from QA knowledge source and 2 from paragraph knowledge source.}
    \label{fig:model}
\end{figure*}
Let $D$ be a QA dataset consisting of $n$ question-answer pairs ($D=\{(q,a)_{i=1}^n\}$), where $q$ and $a$ denote a question and an answer respectively. 
In this work, we are interested in open-domain question answering, in which a model has access to a single or multiple knowledge sources. In this setting, a model conditions on both an input question and additional contexts retrieved from a knowledge source when generating an answer.
\begin{equation*}
    P(a|q,C;\theta) \textnormal{, where } C= \textnormal{top-$k$}(P(\cdot|q;\mathcal{K};\phi))
\end{equation*}
$\theta$ and $\phi$ denote reader and retriever\footnote{this work assumes a dense retriever.} parameters respectively, and $C$ is the $k$ retrieved knowledge records from a knowledge source, $\mathcal{K}$. 


\subsection{Overview}
Our work sheds light on two aspects in open-domain QA: \emph{inference latency} and \emph{extension to multiple knowledge sources with different types.} 
Our model achieves fast inference by incorporating neural memory into the framework. Unlike previous retrieve-and-read approaches, our model does not read retrieved knowledge represented in text. Instead, it conditions on the neural memory representations of the knowledge, retrieved using an off-the-shelf retriever.
Most importantly, our model can incorporate memories from different types of knowledge sources, enriching context information and lifting the restriction on the format of knowledge. Our model is a retrieve-and-read approach with memory, and thus we discuss the reader, retriever, and neural memory in the following sections.


\subsection{Reader}
We utilize the \texttt{T5}-base \citep{t5} encoder-decoder model for a reader.
\subsubsection{Encoder}
\paragraph{Encoding a Question} 
A question is fed to the encoder as in standard practice, yet the core difference is that the question representations are mapped with only the first $j$ layers of the encoder.
\begin{equation}\label{eq:question_map}
    H^q = \texttt{ENC}^{1:j}(t^q(q);\theta_{enc})
\end{equation}
$t^q$ is a template for formatting an input question, i.e. ``Question: \texttt{<q>} Answer:'', which is further described in Appendix \ref{appendix:template}. $\texttt{ENC}^{1:j}$ indicates the first $j$ encoder layers. $H^q$ indicates the $j$-th layer question representations with length $|q|$, $H^q = \{h^q_1, h_2^q, \cdots, h^q_{|q|}\}$, where $|q|$ is equivalent to the length of the question.

\paragraph{Cross Encoding Question and Memories} After obtaining the question representation and top-$k$ memories with a retriever, the encoder cross-encodes the latent representations in the remaining encoder layers to create $\tilde{H}$. 
\begin{equation}\label{eq:cross_encoding}
\begin{aligned}
    \tilde{H} &= \texttt{ENC}^{j+1:L}([M:H^q];\theta_{enc}), 
    \\&\textnormal{where }M = [m^*_1:m^*_2:\cdots:m^*_k]
\end{aligned}
\end{equation}
$[:]$ indicates concatenation and $m^*_i$ indicates a neural memory of varying knowledge formats. We concatenate the retrieved memories $M$ and the latent question representation $H^q$ on the $j$-th layer, and the remaining encoder layers cross-encode the representations. $\tilde{H}$ is the final encoder representations mapped by the remaining layers, from $j+1$ to $L$ encoder layers. 

\subsubsection{Decoder}

The cross-encoded representations hold both a question and retrieved context information. The decoder then creates a prediction based on the fused representations in an auto-regressive fashion. Suppose $\hat{a}_t$ is an answer prediction at time step $t$, then
\begin{equation}
    \hat{a}_t = \texttt{DEC}(\tilde{H}, \hat{a}_{<t};\theta_{dec})
\end{equation}
where $\texttt{DEC}$ is a decoder parameterized with learnable parameters $\theta_{dec}$. 


\subsection{Neural Memory}
Our system uses a neural representation of knowledge, and thus we utilize a model to map raw text knowledge into latent representations. To be specific, we \emph{utilize the same encoder} that is used to encode questions, with the only difference being that a memory is represented with only a fixed length of latent vectors.
\begin{equation*}
    m^* = H^{c^*}_{1:l} \textnormal {, where }H^{c^*} = \texttt{ENC}^{1:j}(t^*(c^*);\theta_{enc})
\end{equation*}
Here $c^*$ denotes knowledge with any type of structure, which is formatted with a knowledge-type-specific template. The first $j$ layers of the encoder are utilized to obtain latent representations of the knowledge, denoted as $H^{c^*}$. Then, the first $l$ vectors of the latent representation are taken as the memory representation of the knowledge. Therefore, knowledge is represented with only $l$ latent vectors, reducing the sequence length. 

Distinct from existing memory-based approaches, such as EMAT \citep{emat}, our framework does not take structure of knowledge into account when mapping to a memory; for instance, \citet{emat} build question and answer memory separately, yet we simply view a QA pair as a single sequence and map it as a single memory. This approach lifts restrictions on the format of knowledge that can be stored as memories. Therefore, our framework can maintain multiple and varying types of knowledge sources, which is hardly feasible with existing methods.
\subsection{Retriever}
Given a question, a retriever maps the question to a single latent representation, and the latent vector is used to search for relevant knowledge in a knowledge pool, stored as (key, value) pairs. Specifically, the search is done in two steps as in Figure \ref{fig:model}: 1) find the index of relevant knowledge snippets through similarity matching of the given question and keys of the knowledge pool with maximum inner product search (MIPS), and 2) retrieve corresponding neural memory knowledge of the top-$k$ indices and pass on the memories for cross-encoding as in Equation \ref{eq:cross_encoding}. The typical item being retrieved in a text-augmented QA model is raw text knowledge \citep{fid,fid_kg, rag,qamat}, whereas our retriever fetches knowledge in the form of a  neural memory. Furthermore, a distinction from existing memory-based methods \citep{emat, qamat} is that our framework utilizes an \textbf{off-the-shelf retriever}, thereby eliminating retriever training which leads to several benefits such as extensiblity which we discuss later in detail. Lastly and most importantly, our retriever retrieves from multiple heterogeneous knowledge sources, one sharp contrast from existing QA models. 




\subsubsection{Training}
\paragraph{Memory Learning Loss}
Our model conditions on retrieved memories during inference for efficiency, and hence, a memory representation is expected to hold salient encoded information of the retrieved knowledge snippet. Therefore, we introduce an auto-encoding loss, $\mathcal{L}_{ae}^*$.
\begin{equation}\label{eq:memory_learning}
\begin{aligned}
    \mathcal{L}_{ae}^* &= -\log P(c^*|m^*;\theta)\\
\end{aligned}
\end{equation}
$\theta$ is a set of learnable parameters of the reader model. The objective is to train the model to reconstruct the original knowledge $c^*$ from its memory representation $m^*$.


\paragraph{Memory-Augmented Generation Loss}
The model is trained to utilize retrieved memories when generating an answer, and thus we define memory-augmented generation loss, $\mathcal{L}_{g}$, as follows.
\begin{equation}
    \mathcal{L}_{g} = -\log P(a|q, M;\theta)
\end{equation}
$M$ is a set of memories retrieved. 

\paragraph{MATTER-QA and MATTER-QA/PRG}
The proposed framework leverages multiple heterogeneous knowledge sources with varying structures. As a result, we propose two model variants: 1) MATTER-QA, the proposed model with QA knowledge source similar to existing memory-based models, and 2) MATTER-QA/PRG, the proposed QA model with multiple heterogeneous knowledge sources, namely QA pairs (QA) and Wikipedia paragraphs (PRG).  

The loss for MATTER-QA is as follows:
\begin{equation}\label{eq:final_qa_loss}
    \mathcal{L}(\theta) = \lambda_g\mathcal{L}_g + \lambda_{ae}\mathcal{L}_{ae}^{qa}
\end{equation}
where $\lambda_i$ is a hyper-parameter to balance the two loss components. 

For MATTER-QA/PRG, an extra loss term is added due to the additional knowledge source.
\begin{equation}
    \mathcal{L}(\theta) = \lambda_g\mathcal{L}_g + \lambda_{ae}^{qa}\mathcal{L}_{ae}^{qa} + \lambda_{ae}^{prg}\mathcal{L}_{ae}^{prg}
\end{equation}
Auto-encoding loss on paragraphs, denoted as $\mathcal{L}_{ae}^{prg}$, is added to Equation \ref{eq:final_qa_loss} to handle the additional unstructured knowledge source. 

\subsection{Discussion on Model Framework}
\paragraph{Shared Encoder}
Utilizing separate encoders for encoding questions and knowledge is a viable option, yet cross-encoding is the core deciding factor why a single encoder is used for both tasks. We take the nature of self-attention into consideration; self-attention is known to attend to similar representations with the use of dot-product. By sharing parameters of the encoder, the latent question representations and related knowledge in neural memories are likely to share similar representations, and hence they are likely to attend to each other during cross-encoding. Empirical results show this to be true as a single encoder approach outperforms separate encoder approach by a meaningful margin.
\paragraph{Attention Complexity}
With the neural memory module, attention complexity significantly drops, leading to a significant boost in speed and less GPU memory\footnote{or CPU memory}.
Numerous studies have found that a decoder takes up majority of time during inference, specifically due to cross-attention \citep{fid,fid-light}; a decoder attends to encoder representations at every time step and at every decoder layer. 
Our approach greatly reduces the length of encoder representations with the introduction of a memory module. For instance, encoder representations with the popular FiD \citep{fid} model have the length of $O(k|c|)$, where $k$ is the number of retrievals and $|c|$ is a length of knowledge. FiD uses 100 Wikipedia documents, with each document consisting of 250 tokens, resulting in an encoded representation of length 25,000. On the other hand, with the same $k$ and memory size set to 2, our approach results in 200 latent vectors in our encoded representation, which is only 0.8\% of tokens used by FiD. The cross-attention has a linear complexity, and thus the time complexity of the proposed model shrinks down proportional to the reduced amount. Lastly, our approach utilizes less GPU memory, as fewer latent vectors from the encoder are stored for computation. The reduced time and space complexity improve the inference speed and GPU usage, respectively. 

\paragraph{Benefits of the Off-the-Shelf Retrieval}
There are several benefits that come from the introduction of an off-the-shelf retriever, of which, the first is faster training. 
Recent retriever augmented QA models \citep{emat,qamat} jointly train retriever and reader models, which require considerable amount of computation and time. Specifically, as training progresses, the knowledge index as a whole is periodically refreshed with new vector representations as the retriever is updated with hard negative sampling. A large knowledge base, such as PAQ \citep{paq}, includes millions of knowledge records. Hence, updating such a large search index introduces significant slowdown in training time. With an off-the-shelf retriever, retriever training and index updates are eliminated, reducing training time and computation cost noticeably.

Furthermore, a plug-and-play approach becomes feasible. With an off-the-shelf retriever, one can switch out different retrievers to trade-off speed and performance based on the use-case at hand. For instance, we show in later sections that our reader model can be combined with a smaller retriever, doubling inference throughput for a small drop in performance.

\paragraph{A Single Retrieval with Multiple Heterogeneous Knowledge Sources}
    As this work deals with multiple knowledge sources with varying structures, the simplest way is to utilize multiple retrievers, one for each knowledge source. However, this approach costs time and resources; multiple retrievers map a question to its own space, and each retriever retrieves from its designated knowledge pool. 
In this work, we mitigate the limitation and utilize a single off-the-shelf retriever model, hence reducing the cost. The intuition is from the recent finding that a well-trained retriever can be used as an universal retriever for varying structures \citep{difar}.
This aspect of ours has largely reduced the inference time and computation cost that otherwise would have been required to maintain multiple retrievers.

\section{Experiment}


\subsection{Model}
MATTER-QA and MATTER-QA/PRG are based on the \texttt{T5-base} model \citep{t5}. Following prior works \citep{qamat, emat}, we train them in two phases: pre-fine-tuning and fine-tuning. During pre-fine-tuning, our models are trained on the PAQ-L1 dataset \citep{paq}, which is a subset of the PAQ dataset consisting of 14.1 million QA pairs. After pre-fine-tuning, the models are further trained on the corresponding downstream dataset. Both memory learning loss and memory-augmented generation loss are used in both pre-fine-tuning and fine-tuning. Model and training hyperparameters are reported in Appendix \ref{appendix:hyperparameter} for reproducibility.


\subsection{Knowledge Source}

For our experiment, we employ two knowledge sources:
\paragraph{Semi-Structured Knowledge (QA Pairs)} We use the PAQ dataset, which contains 64.9 million question-and-answer pairs \citep{paq}.

\paragraph{Unstructured Knowledge (Plain Text)}     We leverage Wikipedia paragraphs, the same set used in previous works\footnote{\url{https://dl.fbaipublicfiles.com/dpr/wikipedia_split/psgs_w100.tsv.gz}} \citep{fid, dpr, fido}. This unstructured knowledge pool comprises 21 million paragraphs, with details provided in Appendix \ref{appendix:dataset_and_knowledge_base}.





\subsection{Dataset}
We test the proposed framework on the three popular open-domain QA datasets, namely TriviaQA \citep{tqa}, Natural Questions \citep{nq}, and Web Questions \citep{wq}. 


\subsection{Baselines \& Metrics}
We compare our proposed approach with four categories of QA models: closed-book, retriever-only, retrieve-and-read, and memory-based QA models. In the closed-book category, we utilize variants of the \texttt{T5} model. For retriever-only models, we compare with RePAQ models of varying sizes \citep{paq} and a retriever coupled with a a reranker. In the retrieve-and-read category, we consider RAG \citep{rag}, FiD \citep{fid}, and QAMAT \citep{qamat}, which are arguably the most popular approaches. Finally, for models with neural memory, we compare with EMAT \citep{emat}. For models that require index search, we have used \texttt{faiss} library with HNSW index as in \citep{paq}. We have run all of the models on a single V100 machine.

To measure QA performance, we report Exact Match (EM). We also report the number of questions processed per second to measure inference speed, which helps assess throughput. For models that retrieve from a knowledge source, we provide details about the type of knowledge source and the number of retrievals ($k$). Additionally, we report model parameter count and CPU RAM usage for memory-based approaches.


\subsection{Experiment Result}

\begin{table*}
    \centering
    \resizebox{\textwidth}{!}{%
    \begin{tabular}{c|c:c:cc:ccc:c:cc:c}
    \hline
    \hline
    \multirow{2}{*}{Setting} & \multirow{2}{*}{Type}& \multirow{2}{*}{Model} & \multicolumn{2}{c}{Knowledge Source} & \multicolumn{3}{c}{Benchmark}& \multicolumn{1}{c}{Speed} &\multicolumn{2}{c}{\# Parameter} & CPU RAM \\
    &  & & $k$ &Type& TQA & NQ & WQ& Q/s& Retrieval & Reader & Memory \\
    \hline
    \multirow{5}{*}{Zero-shot}& \multirow{1}{*}{Retrieve-Read}& QAMAT\ding{168}    & 32 &QA& 34.1 & 37.9 & 25.9    & $11^*$ & 110M & 220M     & NA       \\\cline{2-12}
              & \multirow{4}{*}{Memory} & EMAT               & 10 &QA & 32.4 & 30.6 & 25.6    & 190 & Max 110M &220M& 376GB \\\cdashline{3-12}
              & & MATTER-QA (Fast)& 10 &QA &  45.9    &  43.4  &28.0 & 463 & 12M &       220M    & 188GB \\
              & & MATTER-QA         & 10 &QA& 47.5 & 44.4   & 29.7 & 284 &  110M   &220M       & 188GB \\
              & & MATTER-QA/PRG & 10 &QA/PRG& \textbf{51.6} & \textbf{45.8} & \textbf{29.9}    & 176 & 110M &220M         & 334GB \\
              \hline\hline
    \multirow{16}{*}{Fine-tuned} & \multirow{4}{*}{Closed-book}& T5-base             & NA &NA  & 23.8    & 25.9 & 27.9    & 807 & NA &220M                              & NA    \\
             & & T5-large            & NA  & NA & 28.7    & 28.5 & 30.6    &             289       & NA&770M                      & NA\\
             & & T5-3B               & NA  & NA & 33.6    & 30.4 & 33.6    &           86    &NA& 3B                       & NA    \\
             & & T5-11B              & NA   & NA& 42.3    & 32.6 & 37.2   &           --         & NA&11B                    & NA    \\\cline{2-12}
            &  \multirow{4}{*}{Retrieve}    & RePAQ-256 & --  &QA& 40.2  & 41.4 & -- &     1135          & 12M &     NA                              & NA  \\
             &    & RePAQ-Base          & --  &QA & 39.7    & 40.9 & --    &       418      & 12M &      NA                             & NA  \\
             &     & RePAQ-Xlarge        & --  &QA&   41.3  & 41.7 & -- &      156          & 60M & NA & NA  \\
             &     & RePAQ+Reranker        & 50  &QA&  51.2   & 47.4 &  -- &             50 &  24M& NA & NA  \\\cline{2-12}
             & \multirow{4}{*}{Retrieve-Read} & RAG                 & 10 &PRG& 56.8    & 44.5 & 45.2    &       22             & 110M &400M    &  NA     \\
             &  & FiD-Base            & 100 &PRG & 65      & 48.2 & 32.4    &          2.2            & 110M & 220M            &   NA    \\
             &  & QAMAT         & 32 &QA & 48      & 44.7 & 39.4    &   $11^*$                 & 110M & 220M     & NA       \\
             & & QAMAT\ding{168}     & 32 &QA & 53.2    & 44.5 & 43      &      $11^*$              & 110M & 220M     &    NA   \\\cline{2-12}
             &\multirow{4}{*}{Memory}            & EMAT                & 10 &QA & 44.4    & 44.3 & 36.7    &       190             & Max 110M &220M & 376GB \\\cdashline{3-12}
             & & MATTER-QA (Fast)& 10 &QA &   49.3    &  43.6  &38.0 & 463 & 12M &       220M    & 188GB \\
             & & MATTER-QA         & 10 &QA & 51.2      & 44.8   & 39.2 & 284 & 110M & 220M          & 188GB \\
             & & MATTER-QA/PRG  & 10  &QA/PRG& 56.0    & 46.5 & 40.6    &              176      & 110M&220M          & 334GB\\
    \hline
    \hline
        \end{tabular}}
    \caption{Exact match (EM) score on the three QA benchmarks. QAMAT\ding{168} indicates that the model uses an additional QA set, additional to PAQ dataset. * indicates that QAMAT's inference speed is computed with relative speed compared to that of FiD as reported in the original paper.}\label{tab:main}
    \end{table*}
We evaluate our approach in two settings: zero-shot (after pre-fine-tuning) and fine-tuned (after fine-tuning) stages. The results are shown in Table \ref{tab:main}. In the zero-shot setting, both MATTER-QA and MATTER-QA/PRG significantly outperform QAMAT and EMAT across all datasets. For example, the MATTER-QA model achieves 47.5 EM on TQA in the zero-shot setting, while EMAT scores 32.4 and QAMAT scores 34.1. The performance gap becomes more pronounced with the addition of retrieval from unstructured knowledge pool. When the proposed model conditions on both retrieved QA pairs and Wikipedia paragraphs, the EM score reaches 51.6 on TQA dataset. Both of the proposed models outperform previous approaches by more than 10 EM points. Furthermore, it is worth noting that our approach in the zero-shot setting achieves stronger performance than the fine-tuned baselines of EMAT and QAMAT on TQA and NQ datasets.

In the fine-tuned setting, MATTER achieves competitive performance. The closed-book QA models have low latency but exhibit inferior QA performance. Retrieval-only models demonstrate fast inference speed and competitive EM scores, and coupling them with a reranker can boost performance at the expense of inference speed, presenting a strong baseline. FiD and RAG models exhibit the strongest QA performance across all the datasets; nevertheless, their inference speed is by far the slowest among all the baselines. EMAT, a memory-based approach, overcomes the high latency problem, but the EM scores still have a significant gap compared to those of the retrieve-and-read models. Our models strike a good balance in terms of QA performance and speed trade-off. 
For example, MATTER-QA and MATTER-QA/PRG achieve EM scores of 51.2 and 56.0 respectively on TQA dataset, which are comparable or even better than the scores of RAG model, while being at least 10x faster. Furthermore, MATTER-QA outperforms existing efficient approaches, EMAT and RePAQ, across all datasets in accuracy, while achieving competitive throughput of 284 questions per second. With access to multiple heterogeneous knowledge sources, MATTER-QA/PRG model outperforms MATTER-QA in terms of performance while maintaining a slightly slower, yet competitive throughput.


Our proposed approach offers the added benefit of utilizing an off-the-shelf retriever. Our reader can be easily coupled with a smaller retriever in a plug-and-play fashion, denoted as MATTER-QA (Fast) in Table \ref{tab:main}. The fast version can process 463 questions per second, which is approximately twice as fast as the MATTER-QA model, albeit with a small drop in performance. This clearly demonstrates the flexibility to use different retrievers for varying needs, highlighting one of the core advantages of having an off-the-shelf retriever. Detailed inference speed metrics for each module are provided in Appendix \ref{appendix:inference_speed}.

\section{Analysis}
\paragraph{Retrieval vs Reader in Model Performance}
With the superior results achieved by our proposed model on various QA benchmarks, one natural question arises: are the gains in EM score solely attributable to using an off-the-shelf retriever? In fact, this is not the case; our strong reader models bring such superior EM scores. For instance, the EMAT retriever achieves an EM score of 43.3 on the TQA dataset, while the EM score with the full EMAT model is 44.4, a gain of 1.1 points through reading the retrieved memories. On the other hand, a significant improvement is observed with our reader model: our off-the-shelf retriever model achieves an EM score of 40.0, and when coupled with a reader model that cross-encodes retrieved memories, we achieve a score of 56.0, marking an absolute EM score improvement of 16.0 points. While our retriever is inferior to that of EMAT, the EM score of the full retrieve-read model surpasses that of EMAT by a significant margin. This clearly demonstrates that the retriever is not the core contributor to the performance gains; rather, it is the proposed memory-augmented model that brings a substantial improvement.
\begin{table}[]\label{table:ret_vs_reader}
\centering
\resizebox{\columnwidth}{!}{%
\begin{tabular}{c:ccc:ccc}
\hline
\hline
\multirow{2}{*}{Model}&\multicolumn{3}{c}{TQA}&\multicolumn{3}{c}{NQ}\\
&Ret & Ret+Read& $\Delta$ $(\uparrow)$ &Ret& Ret+Read & $\Delta$ ($\uparrow$)\\
\hline
EMAT & \textbf{43.3} & 44.4& +1.1& 42.2 & 44.3 & +2.1\\
Ours & 40.0 & \textbf{56.0} & \textbf{+16.0} & \textbf{42.5} &\textbf{46.5} & \textbf{+4.0}\\
\hline
\hline
\end{tabular}}
\caption{Comparison between retriever and retrieve-and-read in EM score on TQA and NQ dataset. $\Delta$ denotes the difference in EM score between the retriever and full model (retrieve-and-read).}
\end{table}

\paragraph{Scalability to Varying $k$}
In this section, we analyze the scalability of model performance with varying values of $k$. As shown in Table \ref{tab:varyingK}, in the zero-shot setting, we observe a meaningful improvement by increasing $k$ beyond 10, even though our models are trained with $k$ set to 10. For example, the EM score increases by 0.9 for the QA-only model and 1.9 for the QA/PRG model when $k$ is increased to 20. However, we find that in zero-shot settings, increasing $k$ does not always result in better performance, as evident in the scores when $k$ is set to 30. After fine-tuning, on the other hand, a larger $k$ leads to better QA performance, with the performance linearly increasing with increasing $k$. This demonstrates that our models are capable of answering questions with varying number of retrieved contexts, both in zero-shot settings and after fine-tuning.

Furthermore, we conducted experiments with various combinations of heterogeneous knowledge sources. Our model, MATTER-QA/PRG, has the flexibility to condition on different proportions of these sources. Notably, we observed that our model achieves its highest performance when leveraging knowledge retrieved from both sources simultaneously. While using only a single type of knowledge still outperforms several baselines, it's worth noting that combining information from both sources yields significant improvements over strong baselines, reaching an impressive score of approximately 56 in TriviaQA. This demonstrates the advantage of integrating multiple knowledge sources to enrich the contextual information available to our model.
\begin{table}[t]
\centering
\resizebox{\columnwidth}{!}{%
\begin{tabular}{c:c:cc:ccc:ccc}
\hline
\hline
\multirow{2}{*}{Model}&\multicolumn{3}{c}{Knowledge}&\multicolumn{2}{c}{TriviaQA}\\
&$k$&QA & PRG & Zero-Shot &fine-tuned\\ \hline
QA &10&10 &\xmark&  47.5 &  51.2  \\ 
QA/PRG &10&5&5&   51.6 (+4.1)  &  56.0 (+4.8)  \\ 
\hdashline
QA&20&20&\xmark&  48.6  &    53.1\\ 
QA/PRG&20&10&10&  52.6 (+4.0)  &  58.2 (+5.1) \\ 
\hdashline
QA&30&30&\xmark& 48.2  & 53.4 \\ 
QA/PRG&30&15&15& 52.3 (+4.1)  &  58.7 (+5.3) \\ 
\hline
QA/PRG&10&0&10&  38.6 & 49.3  \\ 
QA/PRG&10&3&7&  51.2 & 56.2  \\ 
QA/PRG&10&7&3&  51.4 &  55.5 \\ 
QA/PRG&10&10&0&  47.7 & 50.9  \\ 
\hline\hline
\end{tabular}}
\caption{EM scores on TriviaQA dataset with varying $k$ and varying proportions. The numbers in the parenthesis are the absolute EM gain by the model with both semi-structured and unstructured knowledge over that of semi-structured only.}\label{tab:varyingK}
\end{table}
\paragraph{A Closer View on Model Conditioning on Heterogeneous Knowledge Base}
We examine what the model focuses on in the retrieved context when generating an answer. In detail, we naively assume that our model has conditioned on specific knowledge when the predicted answer is present within that knowledge. In Table \ref{tab:conditionKB}, we observe that our models employ both retrieved QA pairs and paragraphs during inference. MATTER-QA utilizes both questions and answers in making predictions, while MATTER-QA/PRG equally uses questions, answers, and paragraphs.

\begin{table}[t]
    \centering
    \resizebox{\columnwidth}{!}{%
    \begin{tabular}{c|ccc:ccc}
    \hline
    \hline
    &\multicolumn{3}{c}{QA} & \multicolumn{3}{c}{QA/PRG}\\
    & \% & EM& $\bar{d}(q,\tilde{R})$ & \%& EM & $\bar{d}(q,\tilde{R})$\\
    \hline
    $\hat{a}\in \tilde{Q}$& 4.6\%& 40.4&-- & 0.7\% & 32.4 & --\\
    $\hat{a}\in \tilde{A}$& 62.4\%& 64.8& -- & 13.4\% &51.7&--\\
    $\hat{a}\in \tilde{P}$& -& - & - & 14.3\% & 45.7&--\\
    $\hat{a}\not\in \tilde{R}$& 23.1\%& 7.4 & -- & 18.9\% & 4.1 &--\\
    \hdashline
    $a\in \tilde{R}$& 52.3\%& 84.1&0.37 &  75.1\% &74.0 & 0.36\\
    $a\not\in \tilde{R}$& 47.7\%& 15.2&0.51 & 24.9\% & 1.9 & 0.52\\
    \hline
    \hline
    \end{tabular}}\caption{$\tilde{Q}, \tilde{A}, \tilde{P}$, and $\tilde{R}$ indicate retrieved questions, answers, paragraphs, and the union of retrieved knowledge respectively. In this table, $\in$ indicates that ``is \emph{only} in''. $\bar{d}(q,\tilde{R})$ describes the average distance between question and retrieved knowledge, hence being inverse similarity.}\label{tab:conditionKB}
    \end{table}
An interesting finding is that our models perform exceedingly well when relevant knowledge is provided in the context, i.e., when the answer is present in the retrieved context. When the ground-truth answer is present in the retriever results, our QA model achieves an impressive EM score of 84.7, while our QA/PRG model reaches 74.0. Additionally, our model demonstrates its ability to selectively condition on the retrieved knowledge; the proposed models condition on the retrieved knowledge when the similarity of the retriever results is high. On the other hand, when the retrieved results have low similarity scores to the input question, our model still generates a correct answer that is not present in the retrieved results ~15\% of the time. These key findings reveal that our memory-augmented model can selectively and effectively extract answers from the retrieved knowledge. Moreover, this suggests that our model's performance can be further enhanced when paired with a superior retriever.



\section{Conclusion}
In this paper, we propose an efficient memory-augmented question answering model with multiple heterogeneous knowledge sources. The proposed QA framework is able to retrieve from and condition on retrieved knowledge from multiple sources with varying formats and achieves remarkable performance in popular QA benchmarks, while having high throughput. 

\section*{Limitations}
In this paper, we experiment with two types of knowledge sources, unstructured and semi-structured knowledge sources. Our model can be extended to retrieve from structured knowledge sources, namely knowledge graphs. As knowledge graphs are also a viable source of knowledge, we leave it as a future work to incorporate such knowledge source to enrich the knowledge pool. 

\section*{Ethics Statement}
Most question answering models, including the proposed model, may cause hallucination potentially leading to misinformation. Preventing such issues calls for careful attention and one possible mitigation is to adopt a thresholding approach. In this paper, we demonstrate that an appropriate retriever result is likely to lead to a correct and hallucination-free answer. Combining this insight with the finding that retriever models are well-calibrated \citep{paq}, the confidence score (similarity score) of a retriever can be used as a meaningful proxy for evaluating retrieved results. Hence, one can choose to generate an answer from the reader only if the similarity scores from the retrieval model is above a certain threshold reducing the risk of hallucination.
\bibliography{anthology,custom}

\begin{thebibliography}{22}
\expandafter\ifx\csname natexlab\endcsname\relax\def\natexlab#1{#1}\fi

\bibitem[{Baek et~al.(2023)Baek, Aji, Lehmann, and Hwang}]{difar}
Jinheon Baek, Alham~Fikri Aji, Jens Lehmann, and Sung~Ju Hwang. 2023.
\newblock \href {https://doi.org/10.18653/v1/2023.acl-long.558} {Direct fact retrieval from knowledge graphs without entity linking}.
\newblock In \emph{Proceedings of the 61st Annual Meeting of the Association for Computational Linguistics (Volume 1: Long Papers)}, pages 10038--10055, Toronto, Canada. Association for Computational Linguistics.

\bibitem[{Berant et~al.(2013)Berant, Chou, Frostig, and Liang}]{wq}
Jonathan Berant, Andrew Chou, Roy Frostig, and Percy Liang. 2013.
\newblock \href {https://www.aclweb.org/anthology/D13-1160} {Semantic parsing on {F}reebase from question-answer pairs}.
\newblock In \emph{Proceedings of the 2013 Conference on Empirical Methods in Natural Language Processing}, pages 1533--1544, Seattle, Washington, USA. Association for Computational Linguistics.

\bibitem[{Cai et~al.(2021)Cai, Wang, Li, Lam, and Liu}]{cai-etal-2021-neural}
Deng Cai, Yan Wang, Huayang Li, Wai Lam, and Lemao Liu. 2021.
\newblock \href {https://doi.org/10.18653/v1/2021.acl-long.567} {Neural machine translation with monolingual translation memory}.
\newblock In \emph{Proceedings of the 59th Annual Meeting of the Association for Computational Linguistics and the 11th International Joint Conference on Natural Language Processing (Volume 1: Long Papers)}, pages 7307--7318, Online. Association for Computational Linguistics.

\bibitem[{Chen et~al.(2023)Chen, Verga, de~Jong, Wieting, and Cohen}]{qamat}
Wenhu Chen, Pat Verga, Michiel de~Jong, John Wieting, and William~W. Cohen. 2023.
\newblock \href {https://doi.org/10.18653/v1/2023.eacl-main.117} {Augmenting pre-trained language models with {QA}-memory for open-domain question answering}.
\newblock In \emph{Proceedings of the 17th Conference of the European Chapter of the Association for Computational Linguistics}, pages 1597--1610, Dubrovnik, Croatia. Association for Computational Linguistics.

\bibitem[{de~Jong et~al.(2023)de~Jong, Zemlyanskiy, Ainslie, FitzGerald, Sanghai, Sha, and Cohen}]{fido}
Michiel de~Jong, Yury Zemlyanskiy, Joshua Ainslie, Nicholas FitzGerald, Sumit Sanghai, Fei Sha, and William Cohen. 2023.
\newblock \href {https://doi.org/10.18653/v1/2023.findings-acl.732} {{F}i{DO}: Fusion-in-decoder optimized for stronger performance and faster inference}.
\newblock In \emph{Findings of the Association for Computational Linguistics: ACL 2023}, pages 11534--11547, Toronto, Canada. Association for Computational Linguistics.

\bibitem[{Guan et~al.(2021)Guan, Li, Leng, Lin, Guo, and Zhu}]{DBLP:journals/corr/abs-2112-08560}
Yue Guan, Zhengyi Li, Jingwen Leng, Zhouhan Lin, Minyi Guo, and Yuhao Zhu. 2021.
\newblock \href {http://arxiv.org/abs/2112.08560} {Block-skim: Efficient question answering for transformer}.
\newblock \emph{CoRR}, abs/2112.08560.

\bibitem[{Hofst\"{a}tter et~al.(2023)Hofst\"{a}tter, Chen, Raman, and Zamani}]{fid-light}
Sebastian Hofst\"{a}tter, Jiecao Chen, Karthik Raman, and Hamed Zamani. 2023.
\newblock \href {https://doi.org/10.1145/3539618.3591687} {Fid-light: Efficient and effective retrieval-augmented text generation}.
\newblock In \emph{Proceedings of the 46th International ACM SIGIR Conference on Research and Development in Information Retrieval}, SIGIR '23, page 1437–1447, New York, NY, USA. Association for Computing Machinery.

\bibitem[{Izacard et~al.(2020)Izacard, Petroni, Hosseini, Cao, Riedel, and Grave}]{fid}
Gautier Izacard, Fabio Petroni, Lucas Hosseini, Nicola~De Cao, Sebastian Riedel, and Edouard Grave. 2020.
\newblock \href {https://arxiv.org/abs/2012.15156} {A memory efficient baseline for open domain question answering}.

\bibitem[{{Joshi} et~al.(2017){Joshi}, {Choi}, {Weld}, and {Zettlemoyer}}]{tqa}
Mandar {Joshi}, Eunsol {Choi}, Daniel {Weld}, and Luke {Zettlemoyer}. 2017.
\newblock \href {http://arxiv.org/abs/1705.03551} {{triviaqa: A Large Scale Distantly Supervised Challenge Dataset for Reading Comprehension}}.
\newblock \emph{arXiv e-prints}, page arXiv:1705.03551.

\bibitem[{Karpukhin et~al.(2020)Karpukhin, Oguz, Min, Lewis, Wu, Edunov, Chen, and Yih}]{dpr}
Vladimir Karpukhin, Barlas Oguz, Sewon Min, Patrick Lewis, Ledell Wu, Sergey Edunov, Danqi Chen, and Wen-tau Yih. 2020.
\newblock \href {https://doi.org/10.18653/v1/2020.emnlp-main.550} {Dense passage retrieval for open-domain question answering}.
\newblock In \emph{Proceedings of the 2020 Conference on Empirical Methods in Natural Language Processing (EMNLP)}, pages 6769--6781, Online. Association for Computational Linguistics.

\bibitem[{Khandelwal et~al.(2020)Khandelwal, Levy, Jurafsky, Zettlemoyer, and Lewis}]{knn_lm}
Urvashi Khandelwal, Omer Levy, Dan Jurafsky, Luke Zettlemoyer, and Mike Lewis. 2020.
\newblock \href {https://openreview.net/forum?id=HklBjCEKvH} {Generalization through memorization: Nearest neighbor language models}.
\newblock In \emph{International Conference on Learning Representations}.

\bibitem[{Kwiatkowski et~al.(2019)Kwiatkowski, Palomaki, Redfield, Collins, Parikh, Alberti, Epstein, Polosukhin, Kelcey, Devlin, Lee, Toutanova, Jones, Chang, Dai, Uszkoreit, Le, and Petrov}]{nq}
Tom Kwiatkowski, Jennimaria Palomaki, Olivia Redfield, Michael Collins, Ankur Parikh, Chris Alberti, Danielle Epstein, Illia Polosukhin, Matthew Kelcey, Jacob Devlin, Kenton Lee, Kristina~N. Toutanova, Llion Jones, Ming-Wei Chang, Andrew Dai, Jakob Uszkoreit, Quoc Le, and Slav Petrov. 2019.
\newblock Natural questions: a benchmark for question answering research.
\newblock \emph{Transactions of the Association of Computational Linguistics}.

\bibitem[{Lewis et~al.(2020{\natexlab{a}})Lewis, Liu, Goyal, Ghazvininejad, Mohamed, Levy, Stoyanov, and Zettlemoyer}]{bart}
Mike Lewis, Yinhan Liu, Naman Goyal, Marjan Ghazvininejad, Abdelrahman Mohamed, Omer Levy, Veselin Stoyanov, and Luke Zettlemoyer. 2020{\natexlab{a}}.
\newblock \href {https://doi.org/10.18653/v1/2020.acl-main.703} {{BART}: Denoising sequence-to-sequence pre-training for natural language generation, translation, and comprehension}.
\newblock In \emph{Proceedings of the 58th Annual Meeting of the Association for Computational Linguistics}, pages 7871--7880, Online. Association for Computational Linguistics.

\bibitem[{Lewis et~al.(2020{\natexlab{b}})Lewis, Perez, Piktus, Petroni, Karpukhin, Goyal, K\"{u}ttler, Lewis, Yih, Rockt\"{a}schel, Riedel, and Kiela}]{rag}
Patrick Lewis, Ethan Perez, Aleksandra Piktus, Fabio Petroni, Vladimir Karpukhin, Naman Goyal, Heinrich K\"{u}ttler, Mike Lewis, Wen-tau Yih, Tim Rockt\"{a}schel, Sebastian Riedel, and Douwe Kiela. 2020{\natexlab{b}}.
\newblock \href {https://proceedings.neurips.cc/paper_files/paper/2020/file/6b493230205f780e1bc26945df7481e5-Paper.pdf} {Retrieval-augmented generation for knowledge-intensive nlp tasks}.
\newblock In \emph{Advances in Neural Information Processing Systems}, volume~33, pages 9459--9474. Curran Associates, Inc.

\bibitem[{Lewis et~al.(2021)Lewis, Wu, Liu, Minervini, Küttler, Piktus, Stenetorp, and Riedel}]{paq}
Patrick Lewis, Yuxiang Wu, Linqing Liu, Pasquale Minervini, Heinrich Küttler, Aleksandra Piktus, Pontus Stenetorp, and Sebastian Riedel. 2021.
\newblock \href {http://arxiv.org/abs/2102.07033} {Paq: 65 million probably-asked questions and what you can do with them}.

\bibitem[{Raffel et~al.(2020)Raffel, Shazeer, Roberts, Lee, Narang, Matena, Zhou, Li, and Liu}]{t5}
Colin Raffel, Noam Shazeer, Adam Roberts, Katherine Lee, Sharan Narang, Michael Matena, Yanqi Zhou, Wei Li, and Peter~J. Liu. 2020.
\newblock \href {http://jmlr.org/papers/v21/20-074.html} {Exploring the limits of transfer learning with a unified text-to-text transformer}.
\newblock \emph{Journal of Machine Learning Research}, 21(140):1--67.

\bibitem[{Roberts et~al.(2020)Roberts, Raffel, and Shazeer}]{roberts-etal-2020-much}
Adam Roberts, Colin Raffel, and Noam Shazeer. 2020.
\newblock \href {https://doi.org/10.18653/v1/2020.emnlp-main.437} {How much knowledge can you pack into the parameters of a language model?}
\newblock In \emph{Proceedings of the 2020 Conference on Empirical Methods in Natural Language Processing (EMNLP)}, pages 5418--5426, Online. Association for Computational Linguistics.

\bibitem[{Seonwoo et~al.(2022)Seonwoo, Son, Jin, Lee, Kim, Ha, and Oh}]{two_step}
Yeon Seonwoo, Juhee Son, Jiho Jin, Sang-Woo Lee, Ji-Hoon Kim, Jung-Woo Ha, and Alice Oh. 2022.
\newblock \href {https://doi.org/10.18653/v1/2022.findings-acl.117} {Two-step question retrieval for open-domain {QA}}.
\newblock In \emph{Findings of the Association for Computational Linguistics: ACL 2022}, pages 1487--1492, Dublin, Ireland. Association for Computational Linguistics.

\bibitem[{Song et~al.(2020)Song, Tan, Qin, Lu, and Liu}]{song2020mpnet}
Kaitao Song, Xu~Tan, Tao Qin, Jianfeng Lu, and Tie-Yan Liu. 2020.
\newblock Mpnet: Masked and permuted pre-training for language understanding.
\newblock \emph{arXiv preprint arXiv:2004.09297}.

\bibitem[{Wu et~al.(2020)Wu, Riedel, Minervini, and Stenetorp}]{wu-etal-2020-dont}
Yuxiang Wu, Sebastian Riedel, Pasquale Minervini, and Pontus Stenetorp. 2020.
\newblock \href {https://doi.org/10.18653/v1/2020.emnlp-main.244} {Don{'}t read too much into it: Adaptive computation for open-domain question answering}.
\newblock In \emph{Proceedings of the 2020 Conference on Empirical Methods in Natural Language Processing (EMNLP)}, pages 3029--3039, Online. Association for Computational Linguistics.

\bibitem[{Wu et~al.(2022)Wu, Zhao, Hu, Minervini, Stenetorp, and Riedel}]{emat}
Yuxiang Wu, Yu~Zhao, Baotian Hu, Pasquale Minervini, Pontus Stenetorp, and Sebastian Riedel. 2022.
\newblock \href {https://doi.org/10.18653/v1/2022.emnlp-main.346} {An efficient memory-augmented transformer for knowledge-intensive {NLP} tasks}.
\newblock In \emph{Proceedings of the 2022 Conference on Empirical Methods in Natural Language Processing}, pages 5184--5196, Abu Dhabi, United Arab Emirates. Association for Computational Linguistics.

\bibitem[{Yu et~al.(2022)Yu, Zhu, Fang, Yu, Wang, Xu, Ren, Yang, and Zeng}]{fid_kg}
Donghan Yu, Chenguang Zhu, Yuwei Fang, Wenhao Yu, Shuohang Wang, Yichong Xu, Xiang Ren, Yiming Yang, and Michael Zeng. 2022.
\newblock \href {https://doi.org/10.18653/v1/2022.acl-long.340} {{KG}-{F}i{D}: Infusing knowledge graph in fusion-in-decoder for open-domain question answering}.
\newblock In \emph{Proceedings of the 60th Annual Meeting of the Association for Computational Linguistics (Volume 1: Long Papers)}, pages 4961--4974, Dublin, Ireland. Association for Computational Linguistics.

\end{thebibliography}
\bibliographystyle{acl_natbib}

\appendix

\section{Implementation Details}\label{appendix:hyperparameter}
We set the auto-encoding and memory-augmented generation balancing parameter to 0.3 and 1.0 respectively for both pre-fine-tuning and fine-tuning. We set $k$ to 10 for pre-fine-tuning, 10 QA pairs for MATTER-QA and 5 QA pairs and 5 paragraphs for MATTER-QA/PRG. The learning rates for pre-fine-tuning and fine-tuning are set to 0.0005 and 0.0001 respectively, and we use the linear learning rate decay scheme. We train 10 epochs for pre-fine-tuning and 30 epochs for fine-tuning on TriviaQA and NQ. For WebQuestions, we finetune the models for 50 epochs.  

Our models are initialized with \texttt{t5-base} model. For model-specific hyperparameters, $j$ is set to 8, indicating we use the first 8 encoder layers for both question encoding and memory building. The rest of the encoder layers, 4 encoder layers, are used for cross-encoding memories and a question. Memory size, denoted as $l$, is configured to 2.

For the off-the-shelf retriever, \texttt{mpnet-base} model \citep{song2020mpnet} with mean pooling is used\footnote{\url{https://huggingface.co/sentence-transformers/all-mpnet-base-v2}}. For MATTER-QA (Fast), the off-the-shelf retriever model is RePAQ-256 model \citep{paq}.

For computing EM score, we follow the pre-processing steps used in FiD \citep{fid}, which is specified at the official code repository. For computing the throughput, we perform batch inference and try a max batch size of 500. If a model with the maximum batch size results in the out-of-memory issue, we find the maximum batch size that fits in the memory size. For T5-3B, FiD, and RAG, the batch sizes were set to 300, 15 and 128 respectively. For Reranker, the batch size is set to 200. All the remaining models generate answers with batch size of 500. 
\section{Discussion on Heterogeneous Knowledge Sources}
One natural question with heterogeneous knowledge sources is ``are we giving advantage to our model by giving access to more knowledge, compared to the other baselines?''. The answer is ``No''. The QA pairs, PAQ dataset, are generated from the Wikipedia paragraphs, and hence, the coverage and subject of each knowledge source are the same.

\section{Inference Speed by Each Module}\label{appendix:inference_speed}
Our model performs inference with low latency, and we present the speed of each module of the proposed model with a close view in Table \ref{tab:inference_speed}. We find that a large portion of the inference is spent on the MIPS operation and decoder-side. 
As our model can be switched with an off-the-shelf retriever with a smaller hidden dimension, the total inference time can be greatly reduced; our model with RePAQ-256 retriever runs twice as faster compared to our model with \texttt{mpnet-base} model with reasonable decrease in QA performance as seen in Table \ref{tab:main}.

\begin{table}[t]
\centering
\resizebox{\columnwidth}{!}{%
\begin{tabular}{c:c:c}
\hline
\hline
Module & Process & Latency\\
\hline
\multirow{4}{*}{Memory} & Retriever Forward& 0.012s  \\\cdashline{2-3}
 &Faiss Search - QA (Fast)& 0.41s  \\\cdashline{2-3}
 &Faiss Search - QA& 0.90s  \\\cdashline{2-3}
 &Faiss Search - PRG& 0.56s  \\
 \hline
\multirow{2}{*}{Encoder} & Question Encoding ($\le j$ Layer)& 0.009s  \\
 &Cross-Encoding ($> j$ Layer) & 0.005s  \\
 \hline
 Decoder & Decoder Forward & 0.61s \\
\hline\hline
\end{tabular}}
\caption{Latency report on individual part of the proposed model. $j$ denotes the memory injection layer, and the latency is computed with batch inference, which the batch size is set to 500.}\label{tab:inference_speed}
\end{table}

\section{Preprocessing Knowledge}\label{appendix:dataset_and_knowledge_base}
Our framework maps knowledge into 2 latent representations, and hence the small number of vectors may not fully capture salient information of a long paragraph. In this sense, we split a paragraph into a set, where each item is a two sentence long utterance. 

\section{Templates}\label{appendix:template}
Here, we show the three templates used in the experiments. For input question, the template is as follows:
\begin{quote}
    $t^q(q)$ = Question: $\$q$ Answer:
\end{quote}
$q$ denotes an input question.

For a question-answer pair, the below template is used.
\begin{quote}
    $t^{qa}(q,a)$ = \texttt{<spe1><spe2>} Question: $\$q$ Answer: $\$a$
\end{quote}
\texttt{<spe$i$>} indicates a special token. In this paper, we take the first two representations, and hence we prepend two special tokens in the template for simplicity.

Lastly, for a Wikipedia paragraph, the Wikipedia title and paragraph are used.
\begin{quote}
    $t^{prg}(t, p)$ = \texttt{<spe3><spe4>} Title : $\$t$ Content: $\$p$
\end{quote}
$t$ and $p$ denote Wikipedia title and corresponding content. Note that the special tokens are different from those in the QA template, and we differentiate different knowledge types simply with special tokens.

\section{Case Study}
Table \ref{tab:case_study} shows an example output of MATTER-QA/PRG on a NQ test sample.  
\begin{table*}[h]
    \centering
    \resizebox{\textwidth}{!}{%
        \begin{tabular}{ccc}
        \hline\hline
        Input Question & Answer & Model Prediction\\
        \hline
       \textbf{How many seasons} of the bastard executioner are there? & one season & \textbf{one}\\
       \\
        \hline
        \multicolumn{1}{c}{Retrieved Question} & Retrieved Answer & Relevance \\
        \hline
        \multicolumn{1}{c}{How many episodes of the bastard executioner are there?}& 10 & \xmark\\\hdashline
        \multicolumn{1}{c}{how many episodes are in the bastard executioner?}& 10 &\xmark\\\hdashline
        \multicolumn{1}{c}{how many episodes in the bastard executioner season 1?}& 10 & \xmark\\
        \multicolumn{3}{c}{$\vdots$}\\\\
        \hline
        \multicolumn{2}{c}{Retrieved Passage} & Relevance\\
        \hline
        \multicolumn{2}{c}{\shortstack[c]{On May 22, 2015, ``The Bastard Executioner'' was picked up for a 10-episode series for \\fall launch. On November 18, 2015, FX and Sutter announced that Sutter had canceled the series.}} & \xmark\\\hdashline
        \multicolumn{2}{c}{\shortstack[c]{The Bastard Executioner is an American historical fiction drama \\television series, created by Kurt Sutter and aired on FX from September 15, 2015, to November 17, 2015. \\On November 18, 2015, Sutter announced that FX had \textbf{canceled the series after one season.}}} & \cmark\\\hdashline
        \multicolumn{2}{c}{\shortstack[c]{It ran for 26 episodes, with the last episode airing on 10 January 2016. ``The Executioner'' received \\positive reviews during its broadcast.}} & \xmark\\
        \multicolumn{3}{c}{$\vdots$}\\
        \hline\hline
        \end{tabular}}
        \caption{A NQ test sample output. Relevance indicates if a retrieved knowledge has information to answer the input question.}
        \label{tab:case_study}
    \end{table*}

\end{document}